\pdfoutput=1 
\documentclass[letterpaper, 10 pt, conference]{ieeeconf}  

\IEEEoverridecommandlockouts                              

\usepackage{amsmath} 
\usepackage{amssymb}  
\usepackage{hyperref}
\usepackage{gensymb}
\usepackage{comment}
\usepackage{mathtools}
\usepackage{flushend} 
\usepackage{tikz} 

\usepackage{graphicx}
\graphicspath{{img/}}
\usepackage[caption=false,font=footnotesize]{subfig}

\usepackage{url}
\hypersetup{hidelinks}

\title{\LARGE \bf
Advances in centerline estimation for autonomous lateral control
}

\author{Paolo Cudrano$^{1}$, Simone Mentasti$^{1}$, Matteo Matteucci$^{1}$,\\ Mattia Bersani$^{2}$, Stefano Arrigoni$^{2}$, Federico Cheli$^{2}$ 
\thanks{Partially supported by project TEINVEIN: TEcnologie INnovative per i VEicoli Intelligenti, CUP (Codice Unico Progetto - Unique Project Code): E96D17000110009 - Call "Accordi per la Ricerca e l'Innovazione", cofunded by POR FESR 2014-2020 (Programma Operativo Regionale, Fondo Europeo di Sviluppo Regionale – Regional Operational Programme, European Regional Development Fund).}
\thanks{$^{1}$P. Cudrano, S. Mentasti, M. Matteucci are with the Department of Electronics Information and Bioengineering of Politecnico di Milano, p.zza Leonardo da Vinci 32, Milan, Italy, {\tt\small name.surname@polimi.it}}%
\thanks{$^{2}$M. Bersani, S. Arrigoni, F. Cheli are with the Department of Mechanical Engineering of Politecnico di Milano, via La Masa 1, Milan, Italy, {\tt\small name.surname@polimi.it}}%
}

\newcommand\copyrighttext{%
\footnotesize \copyright~2020 IEEE. Personal use of this material is permitted. Permission from IEEE must be obtained for all other uses, in any current or future media, including reprinting/republishing this material for advertising or promotional purposes, creating new collective works, for resale or redistribution to servers or lists, or reuse of any copyrighted component of this work in other works.} 
\newcommand\copyrightnotice{%
\begin{tikzpicture}[remember picture,overlay]
\node[anchor=south,yshift=10pt] at (current page.south) {\fbox{\parbox{\dimexpr\textwidth-\fboxsep-\fboxrule\relax}{\copyrighttext}}};
\end{tikzpicture}%
}

\begin{document}
\maketitle
\thispagestyle{empty}
\pagestyle{empty}
\copyrightnotice
%
\begin{abstract}
The ability of autonomous vehicles to maintain an accurate trajectory within their road lane is crucial for safe operation. This requires detecting the road lines and estimating the car relative pose within its lane. Lateral lines are usually retrieved from camera images. Still, most of the works on line detection are limited to image mask retrieval and do not provide a usable representation in world coordinates.
What we propose in this paper is a complete perception pipeline based on monocular vision and able to retrieve all the information required by a vehicle lateral control system: road lines equation, centerline, vehicle heading and lateral displacement. We evaluate our system by acquiring data with accurate geometric ground truth. To act as a benchmark for further research, we make this new dataset publicly available at \url{http://airlab.deib.polimi.it/datasets/}.
\end{abstract}

\section{INTRODUCTION}
The ability to drive inside a prescribed road lane, also known as lane following or lane centering, 
is central to the development of fully autonomous vehicles
and involves both perception and control, as it requires to first sense the surrounding environment and then act on the steering accordingly.
\par
Although perception and control are strongly interconnected within this problem, the current literature is partial and fragmented. On the one hand, we have works on lateral control that assume the trajectory to be given and focus on the control models and their implementation~\cite{perez2011cascade}. On the other hand, instead, the perception side is mostly centered on the mere line detection, which is often performed only in image coordinates, so that no line description in the world reference frame is ultimately provided~\cite{kim2008robust}.
\par
To plan the best possible trajectory, however,
it is necessary to retrieve from the environment not only the position and shape of the line markings, but also the shape of the lane center, or centerline, and the vehicle relative pose with respect to it.
This is particularly useful in multi-lane roadways and in
GNNS adverse conditions (e.g. tunnels and urban canyons).
%
\begin{figure}[t!]%
	\centering
	\includegraphics[width=\linewidth]{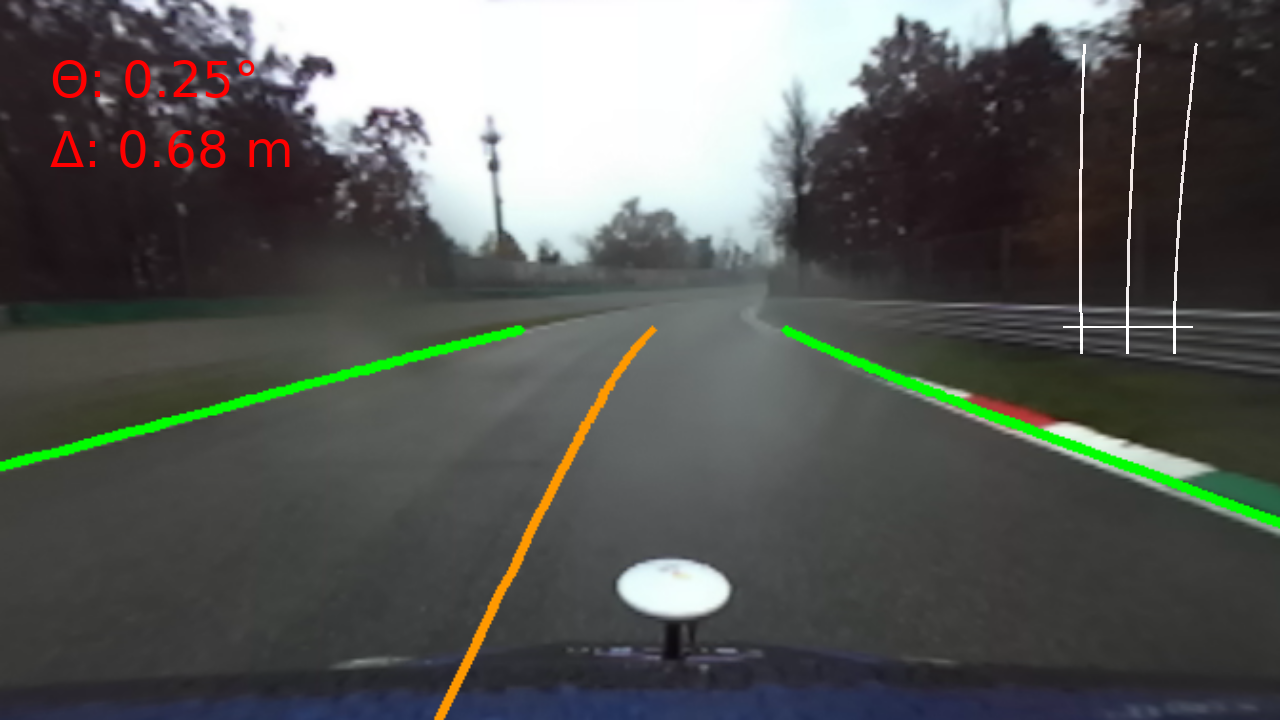}
	\caption{Overview of the output of our system: lateral lines (green), centerline (orange) and relative pose heading and lateral displacement (red). On the right, reconstructed world representation of the scene.}
	\label{fig:todo}
\end{figure}
\par 
At the same time, the technology now commercially available offers only aiding systems. These only
monitor the line position in the strict proximity of the vehicle and are limited to either issue a warning to the drivers (line departure warning systems)~\cite{jeeprenegade2019}, or slightly act on the steering (lane keeping assist) to momentarily adjust the trajectory for them~\cite{fordfusion2020}, although they remain in charge of the vehicle for the entire time~\cite{penmetsa2019potential}.
Only a handful of more advanced commercial systems actually do provide a lane following mechanism, but even in this case it is just for limited situations, such as in presence of a traffic jam (Audi A8 Traffic Jam Pilot~\cite{audia8}), when the vehicle is preceded by another car (Nissan Propilot Assist~\cite{nissanpropilot}), or when driving in limited-access freeways and highways (Autosteer feature in Tesla Autopilot~\cite{teslaautosteer}). 
\par
What we propose in this paper is, 
instead, 
a
complete 
perception 
pipeline 
which enables full lateral control, therefore capable not only to slightly correct the trajectory, but also to plan and maintain it regardless of the particular concurring situations.
To this end,
we design our perception
system 
to provide not only a mathematical description of the road lines in the world frame, but also an estimate of shape and position of the lane centerline and a measurement of the relative pose heading and lateral displacement of the vehicle with respect to it.
\par
The scarcity of similar works in the literature leads also to the absence of related benchmarking data publicly available.
The published datasets in the literature of perception systems only focus on the mere line detection problem~\cite{narote2018review},
even providing no line representation in world coordinates.
In addition, most of these datasets do not contain sequential images~\cite{yu2018bdd100k},~\cite{MVD2017}, or if they do~\cite{pan2018SCNN},~\cite{aly2008real} the sequences are still not long enough to guarantee a fair evaluation of the system performance over time.
No dataset publicly available reports a way to obtain a ground truth measure of the relative pose of the vehicle within the lane, which is crucial for a complete evaluation of our findings.
For this reason, we proceeded to personally collect the data required for the validation of our system 
and we release these data as a further contribution of this work.
\par
To generate an appropriate ground truth and validate our work, the full knowledge of the position of each line marking in the scene was required.
As their measurement is hardly feasible on public roads, 
we performed our experiments on two circuit tracks,
which we could, instead, fully access. 
Although this might seem a simplified environment,
the tracks chosen actually offer a wide variety of driving scenarios and can simulate situations from plain highway driving to conditions even more complicated than usual urban setups,
making the experiments challenging and scientifically significant.
\par
This paper is structured as follows. We first analyze the state of the art concerning line detection, with a particular interest in the models used to represent street lines. Then we proceed with an analysis of the requirements that these systems must satisfy to provide useful information to the control system. Next, in Section IV, we describe our pipeline for line detection and, in Section V, how information like heading and lateral displacement are computed. Lastly, we introduce our dataset and perform an analysis on the accuracy of our algorithms compared to a recorded ground-truth.

\section{RELATED WORK}
\label{sec:related_work}
Lane following has been central to the history of autonomous vehicles.
The first complete work dates back to 1996, when Pomerleau and Jochem~\cite{pomerleau1996rapidly} developed RALPH and evaluated its performance with their test bed vehicle, the Navlab 5, throughout the highways of the United States. Other works, at this early stage, focused mostly on the theoretical aspects, developing new mathematical models for the lateral control problem~\cite{litkouhi1993estimator},~\cite{choi2002lateral}.
\par
At this early stage, the task of line detection began to detach from the rest of the problem~\cite{bertozzi1998gold},~\cite{li2004springrobot}, finding application, later on, within the scope of lane departure warning systems~\cite{bhujbal2015lane}. 
In this context, traditional line detection systems can be generally described in terms of a pipeline of preprocessing, feature extraction, model fitting and line tracking~\cite{narote2018review},~\cite{hillel2014recent},~\cite{zhou2017vision}.
In recent years, learning methods have been introduced into this pipeline. Very common is the use of a Convolutional Neural Network (CNN) as a feature extractor, for its capability of classifying pixels as belonging or not to a line~\cite{chen2018efficient}.
\par
Final output of these systems is a representation of the lines. In this regard, a distinction is made between parametric and non-parametric frameworks~\cite{zhou2017vision}. 
The former
include straight lines~\cite{gaikwad2014lane}, 
used to approximate lines in the vicinity of the vehicle, and second and third degree polynomials, adopted to appropriately model bends~\cite{jung2015efficient}, 
while the latter is mostly represented by non-parametric spline, such as cubic splines~\cite{kim2008robust}, Catmull-Rom splines~\cite{zhao2012novel} and B-snake~\cite{wang2004lane}. While
parametric models provide a compact representation of the curve, as needed for a fast computation of curve-related parameters, non-parametric representations can model more complex scenarios as they do not impose strong constraints on the shape of the road.
\par
As the sole objective of line detection systems is to provide a mathematical description of the lines, any of the described line models is in principle equally valid. For this reason, all of these studies strongly rely on a Cartesian coordinate system as the most intuitive one. In the literature on lateral control instead, the natural parametrization is preferred, as it intrinsically represents the curve shape in terms of quantities directly meaningful for the control model (e.g., heading, radius of curvature, etc.).
In this regard, Hammarstrand et al.~\cite{hammarstrand2016long} argue that models based on arc-length parametrizations are more effective at representing the geometry of a road. Yi et al.~\cite{yi2015vehicle} developed their adaptive cruise control following this same idea and discuss the improvements introduced by a clothoidal model.
\par
Other works in lateral control typically focus on the control models adopted, often 
validating their findings on predefined trajectories.
While this is mostly performed through computer simulations~\cite{yi2015vehicle},~\cite{zhang2009lateral},
P\'erez et al.~\cite{perez2011cascade} make their evaluations on DGPS measurements taken with a real vehicle. Ibaraki et al.~\cite{ibaraki2005design} instead, estimate the position of each line marking detecting the magnetic field of custom markers previously embedded into the lines of their test track.
\par
Only few works incorporate the line detection into their system, aiming at building a complete lane following architecture. In particular, Liu et al.~\cite{liu2007development} first detect the line markings through computer vision and represents them in a Cartesian space, then they reconstruct the intrinsic parameters needed for control. 
To remove this unnecessary complication, Hammarstrand et al.~\cite{hammarstrand2016long} directly represent the detected lines within an intrinsic framework and are able to easily obtain those parameters. 
Their line detection system, however, relies not only on vision to detect the line markings, but also on the use of radar measurements to identify the presence of a guardrail and exploit it to estimate the shape of the road.
\par
In recent years also end-to-end learning approaches have been proposed. Chen and Huang~\cite{chen2017end} developed a CNN-based system able to determine the steering angle to apply to remain on the road. In the meantime, Bojarski et al. \cite{bojarski2016end} present their deep end-to-end module for lateral control, DAVE-2, trained with the images seen by a human driver together with his steering commands, and able to drive the car autonomously for 98\% of the time in relatively brief drives. 
Nonetheless, strong arguments have been raised against the interpretability of end-to-end systems and, ultimately, their safety \cite{koopman2017autonomous}.
\par
Our perception system improves the state of the art as it directly provides the quantities necessary in lateral control, while relying 
on vision and exploiting a compatible road representation. 
Furthermore, an experimental validation is conducted on a real vehicle, considering different scenarios, driving styles and weather conditions.
\section{REQUIREMENTS FOR LATERAL CONTROL}

To properly define a lateral control for an autonomous vehicle, three inputs are essential:
\begin{itemize}
\item vehicle lateral position with respect to centerline;
\item relative orientation with respect to the ideal centerline tangent;
\item roadshape (of the centerline) in front of the vehicle.
\end{itemize}
\begin{figure}[t!]%
	\centering
	\includegraphics[width=\linewidth]{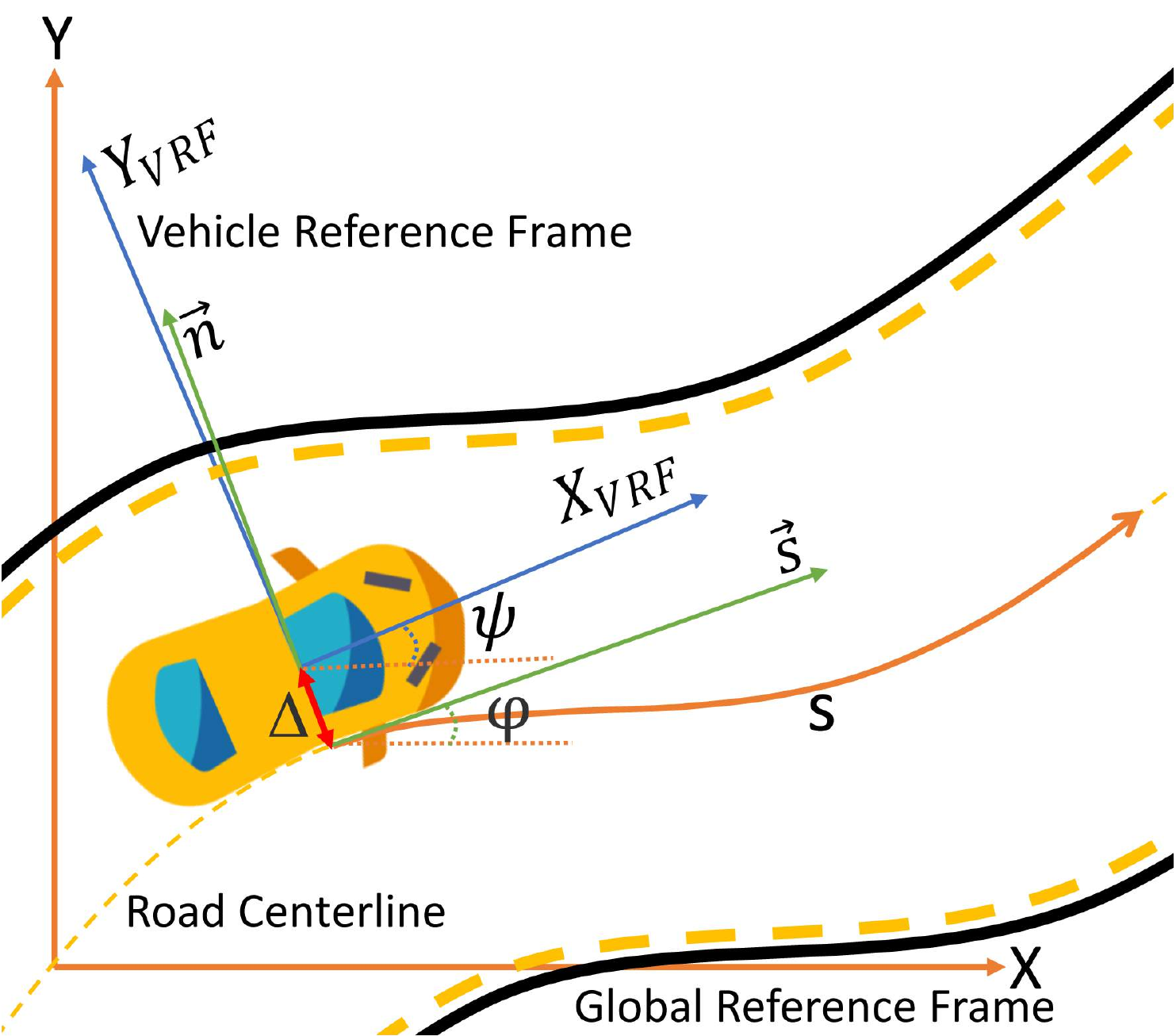}
	\caption{Road representation used in lateral control, highlighting the road centerline as a parametric curve in the vehicle reference frame, and the vehicle relative pose as heading $\Theta = \varphi - \psi$ and lateral displacement $\Delta$.}
	\label{fig:stheta}
\end{figure}
\par
In~\cite{arrigon} the roadshape is described through third order polynomials in a curvilinear abscissa framework $(s-y)$ that is centered according to the current vehicle position. 
The most important advantage with respect to Cartesian coordinates $(X-Y)$ is that each road characteristic can be described as a function of one parameter (i.e., the abscissa \textit{s}), thus each function that approximates the lane center is at least surjective.
This property is very important because it is retained along with the whole optimization horizon in model predictive control approaches~\cite{arrigon}. Fig.~\ref{fig:stheta} depicts an example of such representations.

What is proposed in the following is a pipeline to compute the three parameters required by the control system: vehicle orientation, lateral offset and road shape, in an unknown scenario without the help of GPS data.

\section{LINE DETECTION}
To estimate the required parameters, we need to acquire a representation of the lane lateral lines in the scene. 
\par
At first, we adopt a purpose-built CNN to determine which pixels in an acquired frame belong to a line marking.
Our architecture, trained using the Berkeley DeepDrive Dataset~\cite{berkleydd}, is based on U-net~\cite{unet}, but with some significative changes to improve the network speed on low power devices and allow predictions at 100~Hz. In particular, the depth is reduced to two levels, and the input is downscaled to 512x256 and converted to grayscale. With these changes the network requires only 5ms to predict an image on our testing setup, a Jetson Xavier.
\par
The obtained prediction mask is then post-processed through two stages. At first, we apply an Inverse Perspective Mapping (IPM) and project it into the Bird's Eye View (BEV) space, where the scene is rectified and the shape of the lines reconstructed. In this space, then, the predictions are thresholded and morphologically cleaned to limit the presence of artifacts.
The result is a binary mask in the BEV space, highlighting what we refer to as \textit{feature points}.
\par
Next, a \textit{feature points selection} phase separates the points belonging to each road line of interest, discarding noisy detections at the same time.
Algorithms for connected components extraction and clustering easily fail as soon as the points detected are slightly discontinuous,
and have usually a strong computational demand.
Therefore, we develop for this task a custom algorithm
based on 
the idea that the lateral lines are likely well-positioned at the vehicle sides when looking in its close proximity. Once they are identified there, then, it is easier to progressively follow them as they move further away.
Exploiting this concept, our 
\textit{window-based line following (WLF) algorithm} is able to search for a line in the lower end of the image and then follow it upwards along its shape thanks to a mechanism of moving windows.
\par
The \textit{line points} collected ${P_{l_i} = (x_i, y_i)}_i$ are then passed to the \textit{fitting phase}.
Here each line is first temporarily fit to a cubic spline model to filter out the small noise associated with the detections while still preserving its shape.
This model is however 
too complex and thus
hard to further manipulate.
To obtain a representation useful for lateral control, we propose to represent our line in a curviliear framework ($s - \vartheta$). The conversion 
of the modeled lines into this framework requires a few steps, as the transition is highly nonlinear and cannot be performed analytically. 
We first need to sample our splines, obtaining a set of points $S'_{\left(x,y\right)} = \{ \left(x_i', y_i'\right)\}_{i=0,...,n}$.
Fixing then an origin on the first detected point $(x_0',y_0')$, we measure the euclidean distance $\Delta s_i$ between each point and its successor and the orientation of their connecting segment $\vartheta_i$ with respect to the x axis.
For small increments $\Delta s_i$ then, we can assume:
\begin{equation}
    s_i = \int ds \approx \sum_{k=0}^i \Delta s_k
\end{equation}
obtaining a set $S_{\left(s,\vartheta\right)} = \{ \left(s_i, \vartheta_i\right)\}_{i=1,...,n}$. The main advantage obtained is that this set, while still related to our Cartesian curve, is now representable in the $(s - \vartheta)$-space as a 1-dimensional function:
\begin{equation}
    \vartheta = f(s)
\end{equation}
which can be easily fit with a polynomial model, final representation of our lateral lines.
\par
As last step of our algorithm, the temporal consistency of the estimated lines is enforced in several ways. The information from past estimates is used to facilitate the feature points selection. In particular, when a line is lost because no feature points are found within a window, we can start a recovery procedure that searches for more points in a neighborhood of where the line is expected to be.
\par
Although the algorithm could already properly function,
a further addition is the introduction of the odometry measures, to improve the forward model of the road lines 
and increase the robustness of the system. 
Indeed, 
while we are driving on our lane, we can see its shape for dozens of meters ahead. Thus, instead of forgetting it, we can exploit this information as we move forward, in order to model not only the road ahead of us, but also the tract we just passed. This is crucial to be able to model the road lines not only far ahead of the vehicle, but also and especially where the vehicle currently is. 
To do so, we only need a measurement of the displacement of the vehicle pose between two consecutive frames.
We obtain this information 
from the encoders of the vehicle, which were readily accessible to us. Nevertheless, visual odometry could be used, alternatively,
in order to entirely rely on vision. 
With these
measurements, 
we can store the line points detected at each time step and, at the next step, project them backwards to reflect the motion of our vehicle, finally adding them to the new detections before the line is fitted. As we move forwards, more and more points are accumulated, representing regions of the road not observable anymore. To avoid storing too complex road structures then, we prune old portions of the road as we move away from them, maintaining only past line points within 5--10 meters from our vehicle.
\par
Furthermore, while the literature is mostly oriented towards Bayesian filters (mostly KF and EKF) to track the model parameters, we adopt an alternative perspective. It is important to notice that Bayesian filters directly act on the parameters of the line after the fitting, and for optimal results they require external information about the motion of the vehicle. As our line detection system 
relies instead on vision, we employ 
an adaptive filter based on the Recursive Least Square (RLS) method \cite{campi1994exponentially}. 
In particular, we design this filter to receive as input, at each time step $t$, the set of line points observed in the respective frame. With these, its overall model estimate is updated, following a weighted least squares scheme. Entering the filter with a full weight, points are considered to lose importance as they age, and thus their weight is exponentially reduced over time.
The model we consider is a cubic polynomial in coordinates (${s-\vartheta}$):
\begin{equation}
\vartheta(s) = \mathbf{w}^T \boldsymbol{\phi}(s) = w_3 s^3 + w_2 s^2 + w_1 s + w_0
\end{equation}
where
\begin{align}
\mathbf{w} &= \begin{pmatrix} w_3 & w_2 & w_1 & w_0 \end{pmatrix}^T \\
\boldsymbol{\phi}(s) &= \begin{pmatrix} s^3 & s^2 & s & 1 \end{pmatrix}^T
\end{align}
At a given time $t$, we observe the points $\{ \left(s_{t,i}, \vartheta_{t,i}\right)\}_{i=1,...,n}$, and we can then build:
\begin{equation}
\Phi_t = \begin{pmatrix} 
\boldsymbol{\phi}(s_{t,1})^T \\
\vdots \\
\boldsymbol{\phi}(s_{t,n})^T
\end{pmatrix} = \begin{pmatrix} 
s_{t,1}^3 & s_{t,1}^2 & s_{t,1} & 1 \\
\vdots & \vdots & \vdots & \vdots \\
s_{t,n}^3 & s_{t,n}^2 & s_{t,n} & 1 \\
\end{pmatrix}
\end{equation}
We consider our measurements to be constituted of a deterministic term, to be estimated, and a stochastic term, to be removed:
\begin{equation}
\boldsymbol{\vartheta}_t = \Phi_t \mathbf{w} + \boldsymbol{\eta}_t, \qquad \boldsymbol{\eta}_t \sim \mathcal{N}(\mathbf{0},\,\Sigma)
\end{equation}
With this in mind then, at a given time $t$, we can update our model $\mathbf{w}$ with a forgetting factor $\mu$ by computing, for each $i=1,\dots,n$:
\begin{align}
\tilde{\mu} &= \begin{cases} 
\mu & \text{if } i=1 \\
1 & \text{otherwise}
\end{cases}\\
e_{t,i} &= \vartheta_{t,i} - \mathbf{w}^T \boldsymbol{\phi}(s_{t,i}) \\
\tilde{R} &= \left( 1 + \frac{1}{\tilde{\mu}}\, \boldsymbol{\phi}(s_{t,i})^T \, R\, \boldsymbol{\phi}(s_{t,i}) \right)^{-1} \\
R &= \frac{1}{\tilde{\mu}} \left( R - \frac{1}{\tilde{\mu}}\, R \, \boldsymbol{\phi}(s_{t,i})\, \tilde{R} \, \boldsymbol{\phi}(s_{t,i})^T\, R \right) \\
\mathbf{\Delta w} &= e_{t_i} \cdot R \, \boldsymbol{\phi}(s_{t,i}) \\
\mathbf{w} &= \mathbf{w} + \mathbf{\Delta w}
\end{align}
where $R$ and $\mathbf{w}$ are updated at each step.
The main advantage of this approach is that no assumption is made on the behavior of the parameters and it is instead only the accumulation of line points through time to smooth the results.
The recursive formulation then makes the computation fast and efficient.

\section{CENTERLINE AND RELATIVE POSE ESTIMATION}
Given the representation of the lateral lines, it is important to model the lane centerline and the relative pose of the vehicle, measured in terms of its heading $\Theta$ and lateral offset $\Delta$ with respect to the centerline.
\par
As no parallelism is enforced between the lateral lines, an analytical representation of the centerline is hard to find, 
but we can reconstruct its shape with some geometrical consideration and exploiting the 
line 
representation adopted.
In particular, we devise an algorithm 
to project the points from both lateral lines into the same $(s - \vartheta)$ plane, 
and we fit these points with a single model, equally influenced by both lines.
In the best scenario, this would require each line point to be projected towards the center, along the normal direction to the road. This projection, particularly impractical in Cartesian coordinates, is easily achieved 
in the space $(s - \vartheta)$.
\par
We assume, for the time being, that the lane has a fixed curvature. Moreover, if we take into account the center of curvature $\boldsymbol{C}$ of any road line, we also make the following assumptions:
\begin{itemize}
	\item the two lateral lines ($l_l$,$l_r$) and the centerline ($l_c$) share the same center of curvature ($ \boldsymbol{C} $):
	\begin{equation}
		\boldsymbol{C}_{l_l} \equiv \boldsymbol{C}_{l_r} \equiv \boldsymbol{C}_{l_c} \equiv \boldsymbol{C}
	\end{equation}
	\item the center of curvature varies smoothly:
	\begin{equation}
		\boldsymbol{C}^t \approx \boldsymbol{C}^{t-1}
	\end{equation}
\end{itemize}
\par
\begin{figure}[t!]%
	\centering
	\includegraphics[width=\linewidth]{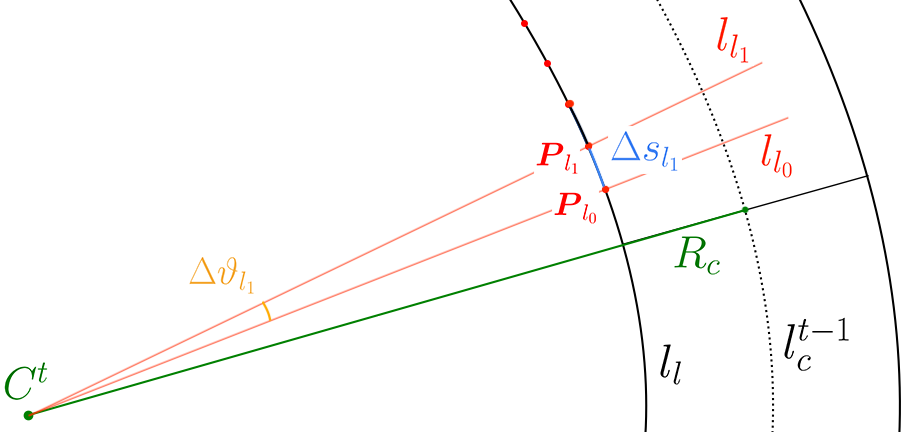}
	\caption{
    Representation of the geometrical transformation performed for the centerline estimation, highlighting each quantity involved.
    }
	\label{fig:algorithm_centerline}
\end{figure}

 
With this setup, we can define the following procedure, to be repeated for both lateral lines (generically indicated as $l$). We refer the reader to Fig.~\ref{fig:algorithm_centerline} for a graphical interpretation of the quantities involved.
\begin{enumerate} 
	\item Compute $\boldsymbol{C}^t$ from $l_c^{t-1}$, using its heading and radius of curvature. 
	Define the orthogonal to $l_c^{t-1}$ as $l_{lc}$ (in green). 
	\item Find the line $l_{l_0}$ passing through $\boldsymbol{C}^t$ and $\boldsymbol{P_{l_0}}$, the first line point in $l$.
	\item Find the line $l_{l_1}$ passing through $\boldsymbol{C}^t$ and $\boldsymbol{P_{l_1}}$, the second line point in $l$.
	\item Compute $ R_l = ||\boldsymbol{P_{l_0}} - \boldsymbol{C}^t ||_2 $.
	\item Compute the angle $ \Delta \vartheta_{l_1}$ between $l_{l_0}$ and $l_{l_1}$.
	\item Compute $ \Delta s_{l_1} = R_l \cdot \Delta \vartheta_{l_1} $.
	\item Compute $ \Delta s_{c_1} = R_c \cdot \Delta \vartheta_{l_1} $.
	\item Obtain the ratio $ r_{s_l} = \frac{\Delta s_{l_1}}{\Delta s_{c_1}} $.
	\item Compute the angle $ \Delta \vartheta_{l_0}$ between $l_{lc}$ and $l_{l_0}$.
	\item Define, for convenience, 
	\begin{equation}\label{eq:4_3_1_1_curvature_ratio}
		s_{c_0} = R_c \cdot \Delta \vartheta_{l_0}
	\end{equation}
\end{enumerate}
\par
At this point, with the ratio in Equation~\eqref{eq:4_3_1_1_curvature_ratio}, we can define a coordinate transformation
from the lateral line to the centerline and vice versa, all remaining into the parametric framework.
For any line point $\left(s_{l_i}, \vartheta_{l_i}\right)$:
\begin{equation}
\begin{array}{rcl}
s_{c_i} &=& s_{c_0} + r_{s_l} \left( s_{l_i} - s_{l_0} \right) = s_{c_0} + r_{s_l} \cdot s_{l_i} \\
\vartheta_{c_i} &=& \vartheta_{l_i}
\end{array}
\end{equation}
Notice that, although we think of this projection in the Cartesian space, 
we only define a linear transformation in the $s-\vartheta$ frame, aiming at rescaling each line model in order to make their shapes comparable.
Although the assumptions made do not hold in general scenarios, 
this produces an acceptable approximation of the expected results.
\par
With this procedure then, we are ultimately able to take all the points detected on both lines and collapse them onto the centerline. \par
As done for the lateral lines, the points can be fit with a cubic polynomial in $(s - \vartheta)$ and the result tracked through an RLS framework.

\subsection{Relative pose estimation}

\begin{figure}[t!]%
	\centering
	\includegraphics[trim={1.6cm 0.3cm 0 1.55cm},clip, width=\linewidth]{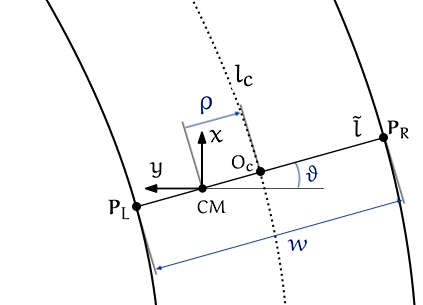}
	\caption{
	Representation used in the vehicle relative pose estimation, including state and observation variables for the EKF filter adopted.}
	\label{fig:ekf_param}
\end{figure}

Given the centerline model, we notice that the heading of the vehicle is represented by the value of $\vartheta(s)$ in a particular point, to be determined. Finding the exact point however is not simple, as we want to perform this measurements exactly along the line passing through the center of mass of the vehicle $CM$. 
As this requires us to pass from intrinsic to extrinsic coordinates, no closed form formulas are available, and we have to solve a simple nonlinear equation. In particular, as illustrated in Fig.~\ref{fig:ekf_param}, we need to look for a line $\tilde{l}$, passing through $CM$ and crossing the centerline $l_c$ perpendicularly. Formally, we search for a value $\tilde{s}$, corresponding to a point $O_c$ along the centerline, such that:
\begin{equation}
\begin{cases}
	CM \in \tilde{l} \\
	O_c \in \tilde{l} \\
	\tilde{l} \perp l_c
\end{cases}
\end{equation}
This can be easily done translating this conditions in the corresponding geometrical equations and considering that, for parametric representations:
\begin{equation}
x = \int cos(\vartheta(s)) ds, \quad y = \int sin(\vartheta(s)) ds
\end{equation}
\par
Once this point is found, heading ($\Theta$) and lateral displacement ($\Delta$) are:
\begin{align}
\Theta&= \vartheta(\tilde{s}) \\
\Delta &=  \begin{cases}
+ || O_c - CM ||_2 & \mbox{if } O_{c_y} \ge 0 \\
- || O_c - CM ||_2 & \mbox{if } O_{c_y} < 0 \\
\end{cases}
\end{align}
\par
To maintain the temporal consistency of the vehicle pose, we set up an Extended Kalman Filter (EKF). We take as measurements the Cartesian position of the points $\boldsymbol{P}_L$ and $\boldsymbol{P}_R$, intersections of the lateral lines with $\tilde{l}$ (see  Fig.~\ref{fig:ekf_param}), and maintain a state composed of $\vartheta$, heading of the vehicle relative to the centerline, $\rho$, signed normalized lateral displacement, and $w$, width of the road. 
Notice that we formally split the lateral displacement $\Delta$ into $w$ and $\rho$, measuring respectively the width of the lane and the relative (percentage) offset with respect to it. This is done on one hand to simplify the definition of the measurement function and thus obtain faster convergence, and on the other hand to obtain the additional estimate of $w$, potentially helpful in control.
This allows us to produce approximate estimates even when the tracking for one of the two lateral lines is lost, as we can locally impose parallelism and project the detected line on the opposite side of the lane, allowing our system to be resilient for short periods of time.
\par
Mathematically, the state space model representation of our system can be shown as:
\begin{align}
\boldsymbol{x} &= \begin{pmatrix}
\vartheta \\
\rho \\
w
\end{pmatrix}
, \qquad
\boldsymbol{z} = \begin{pmatrix}
x_{P_L} \\
y_{P_L} \\
x_{P_R} \\
y_{P_R} \\
\end{pmatrix}
\\*
\boldsymbol{x}^t &= \boldsymbol{x}^{t-1} + w_t , \qquad w_t \sim \mathcal{N}(0,\,Q)\,. \\*
\boldsymbol{z}^t &= h(\boldsymbol{x}^t) + v_t , \qquad v_t \sim \mathcal{N}(0,\,R)\,.
\end{align}
where the measurement function $h$ is:
\begin{equation}
h(\boldsymbol{x}) = \begin{pmatrix}
x_{CM} - \frac{w}{2} (1-\rho) \sin\vartheta \\
y_{CM} + \frac{w}{2} (1-\rho) \cos\vartheta \\
x_{CM} + \frac{w}{2} (1+\rho) \sin\vartheta \\
y_{CM} - \frac{w}{2} (1+\rho) \cos\vartheta
\end{pmatrix},
\end{equation}
with $CM = ( x_{CM}, y_{CM} ) $ center of mass of the vehicle.

\section{EXPERIMENTAL VALIDATION}
The dataset used for our tests is made publicly available at \url{http://airlab.deib.polimi.it/datasets/}.
\par
\begin{figure}[t!]%
\centering
\subfloat[Track Aci-Sara Lainate\label{1a}]{
    \includegraphics[width=\linewidth]{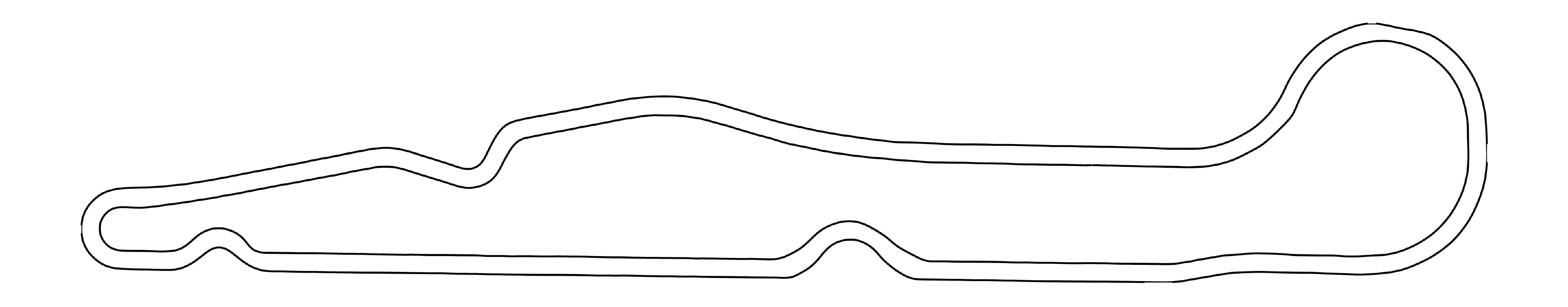}
}\\[5px]
\subfloat[Monza Eni Circuit\label{1b}]{
    \includegraphics[width=\linewidth]{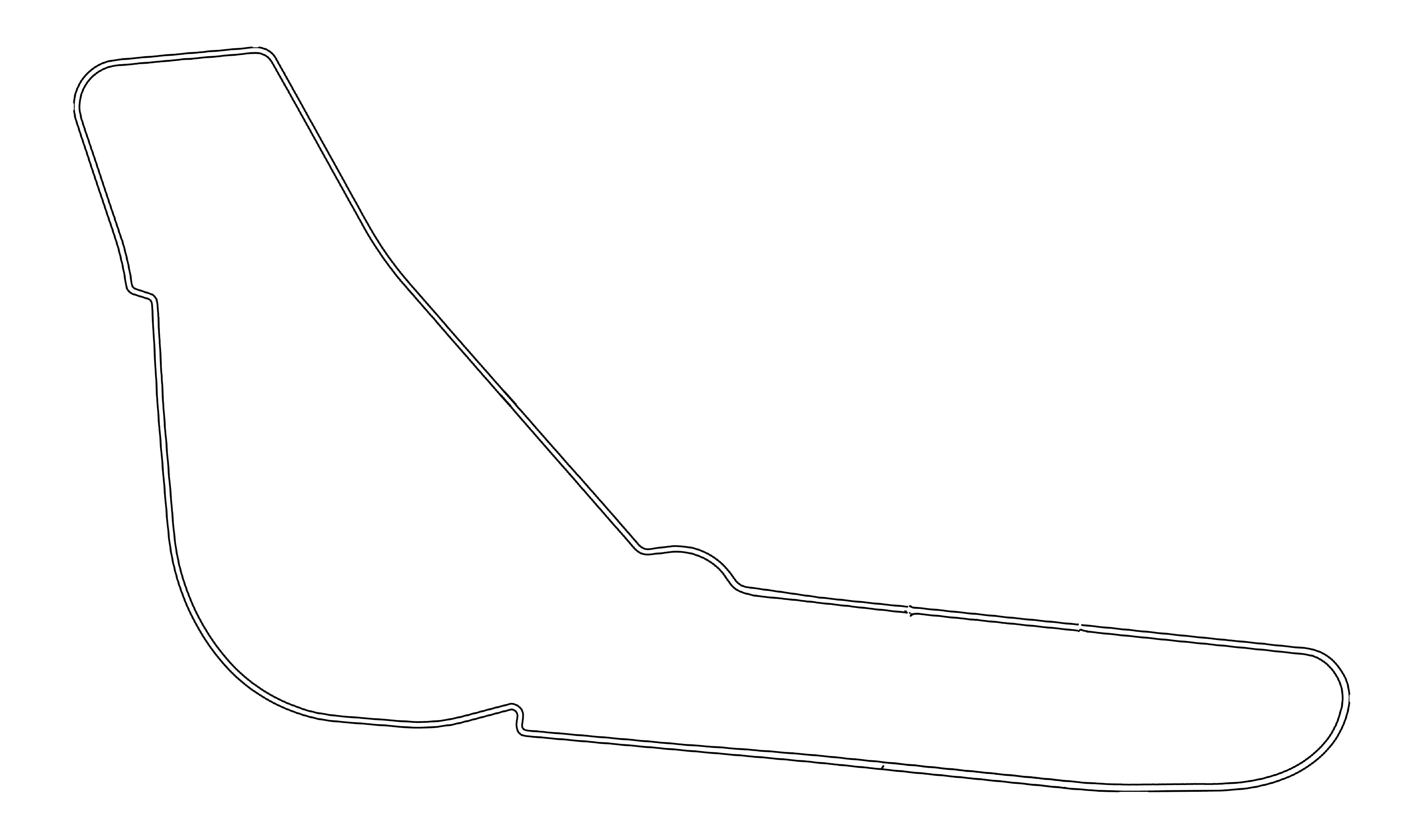}
}
\caption{The two racetracks used to collect our dataset.}
\label{fig:tracks}
\end{figure}
\begin{figure}[t!]%
\centering
	\includegraphics[ width=\linewidth]{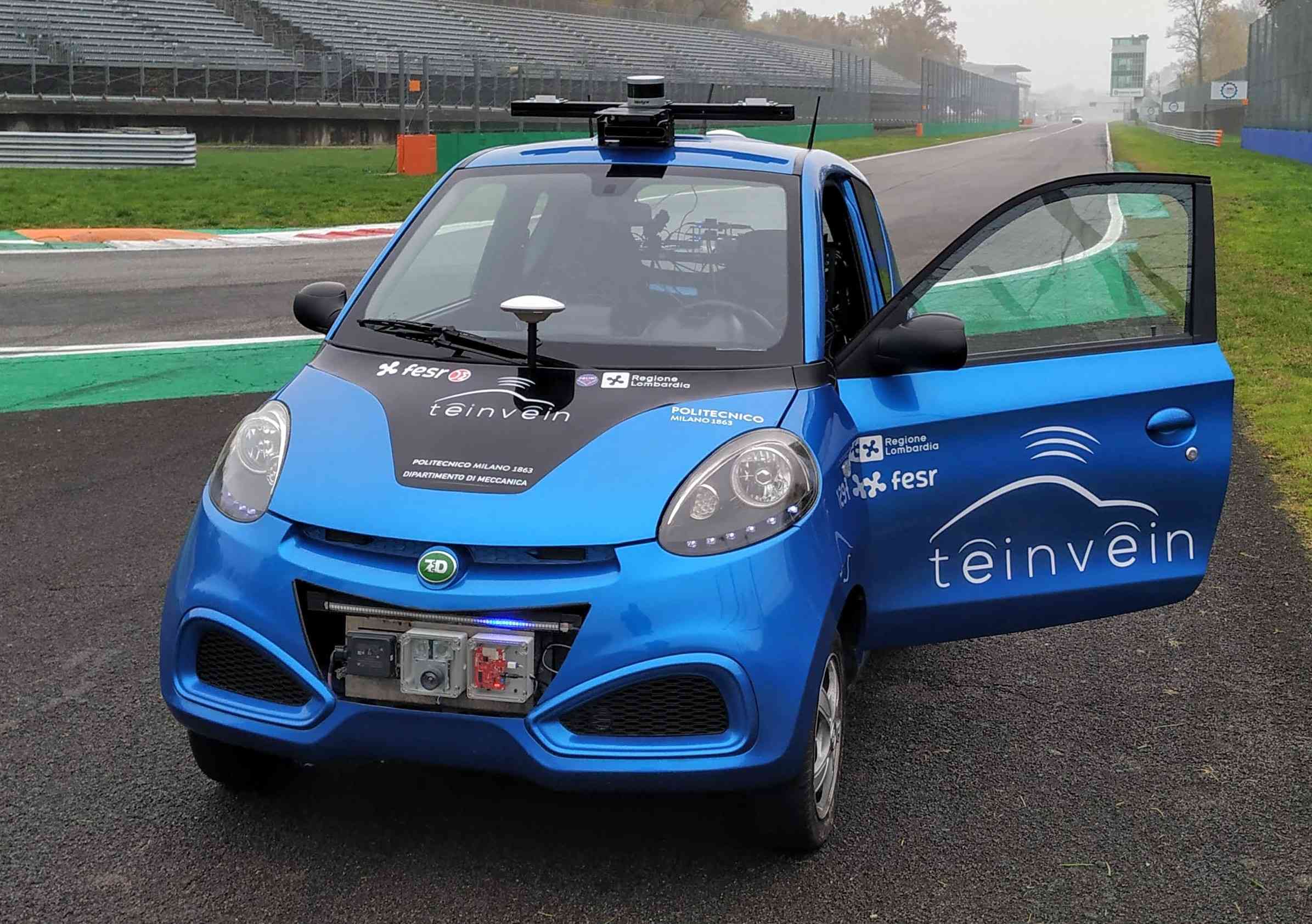}
    \caption{
	Image of the experimental vehicle used for the dataset acquisition.}
	\label{fig:car}
\end{figure}

All data have been acquired on the Aci-Sara Lainate~(IT) racetrack and on the Monza Eni Circuit track (Fig.~\ref{fig:tracks}).
The two circuits present an optimal configuration for real street testing with long straights, ample radius curves and narrow chicanes.
The dataset is acquired using a fully instrumented vehicle, shown in Fig.~\ref{fig:car}. Images with resolution $672\,\times\,376$ are recorded using a ZED stereo-camera working at $100\,Hz$. Car trajectory and lines coordinates are registered using a Swiftnav RTK GPS.
\par
The ground truth creation process requires to map the desired area and retrieve the lines GPS coordinates.
Then, the road centerline has been calculated considering the mean value of the track boundaries and sampled to guarantee a point each $ ds = 0.5 \,m$. This value of $ds$ allows avoiding the oversampling of GPS signals while ensuring smoothness and accuracy of the road map. After that, third order polynomials have been derived at every $ds$ along the centerline for the following $30$ meters. 
Thanks to the experimental data collected, the lateral distance from the centerline is computed as the minimum distance to the closest point of the centerline map. The relative angle with respect to the centerline is instead evaluated by approximating the centerline orientation computing the tangent to the GPS data.

\begin{figure}[t!]%
\centering
\subfloat[Track A - Centered\label{a1}]{
    \includegraphics[width=0.47\linewidth]{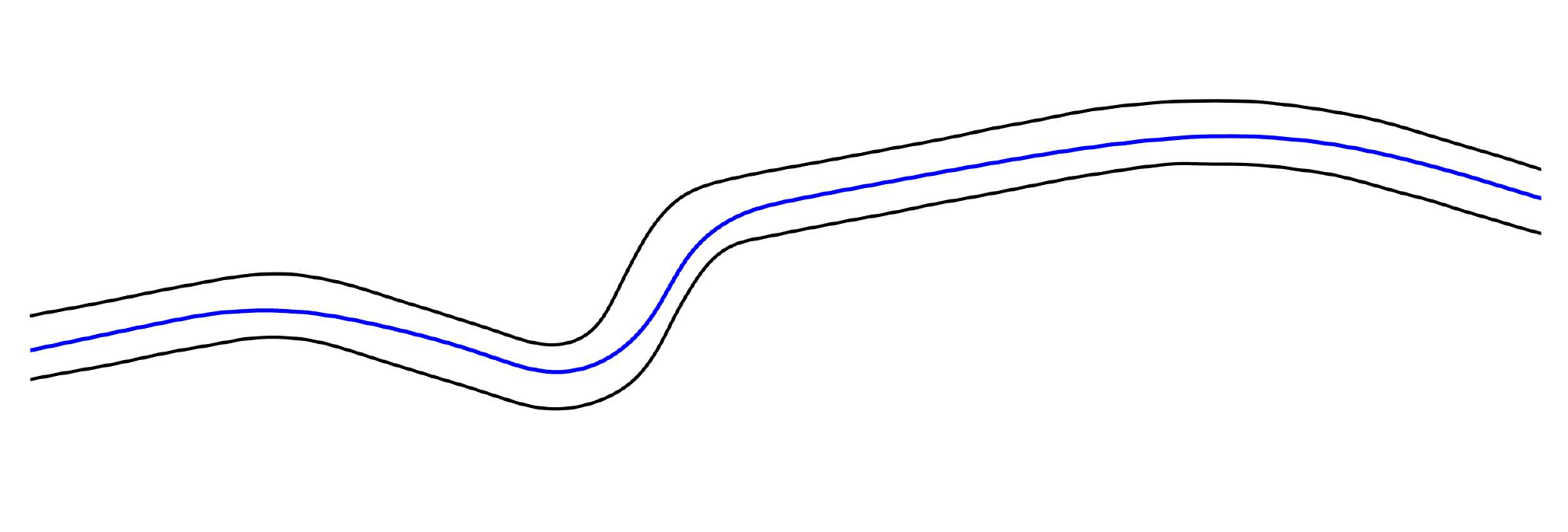}
}\hfill
\subfloat[Track A - Oscillating\label{a2}]{
    \includegraphics[width=0.47\linewidth]{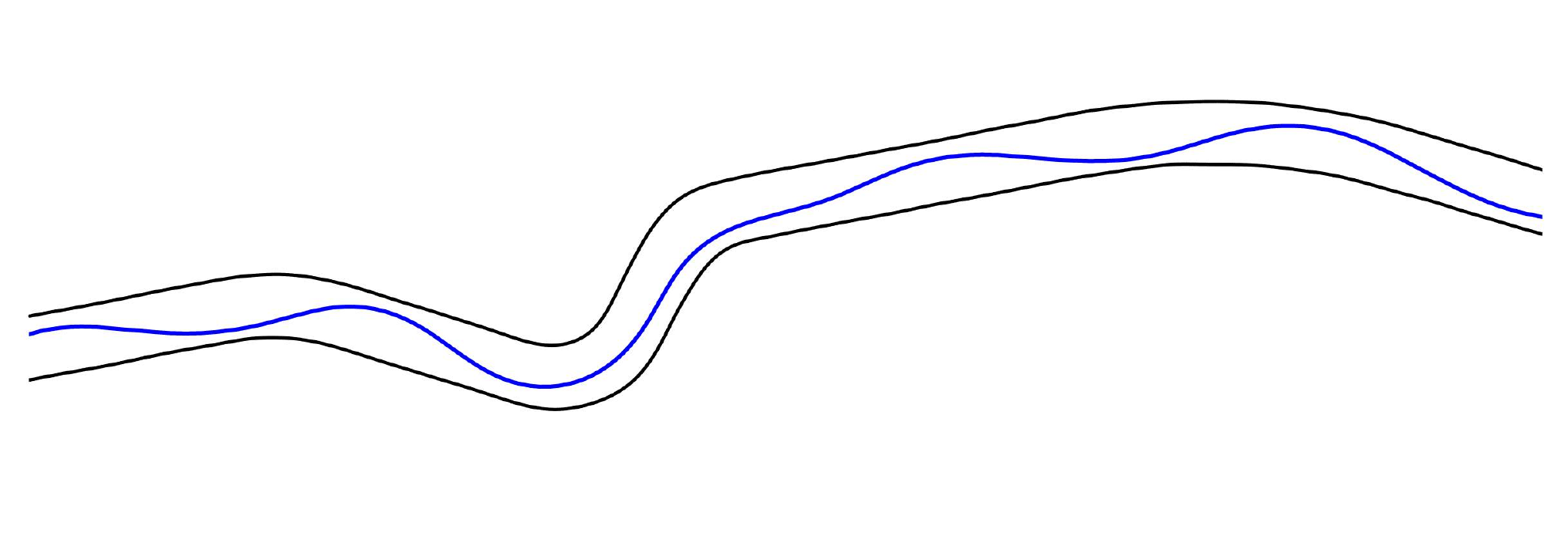}
}\\
\subfloat[Track B - Centered\label{b1}]{
    \includegraphics[width=0.47\linewidth]{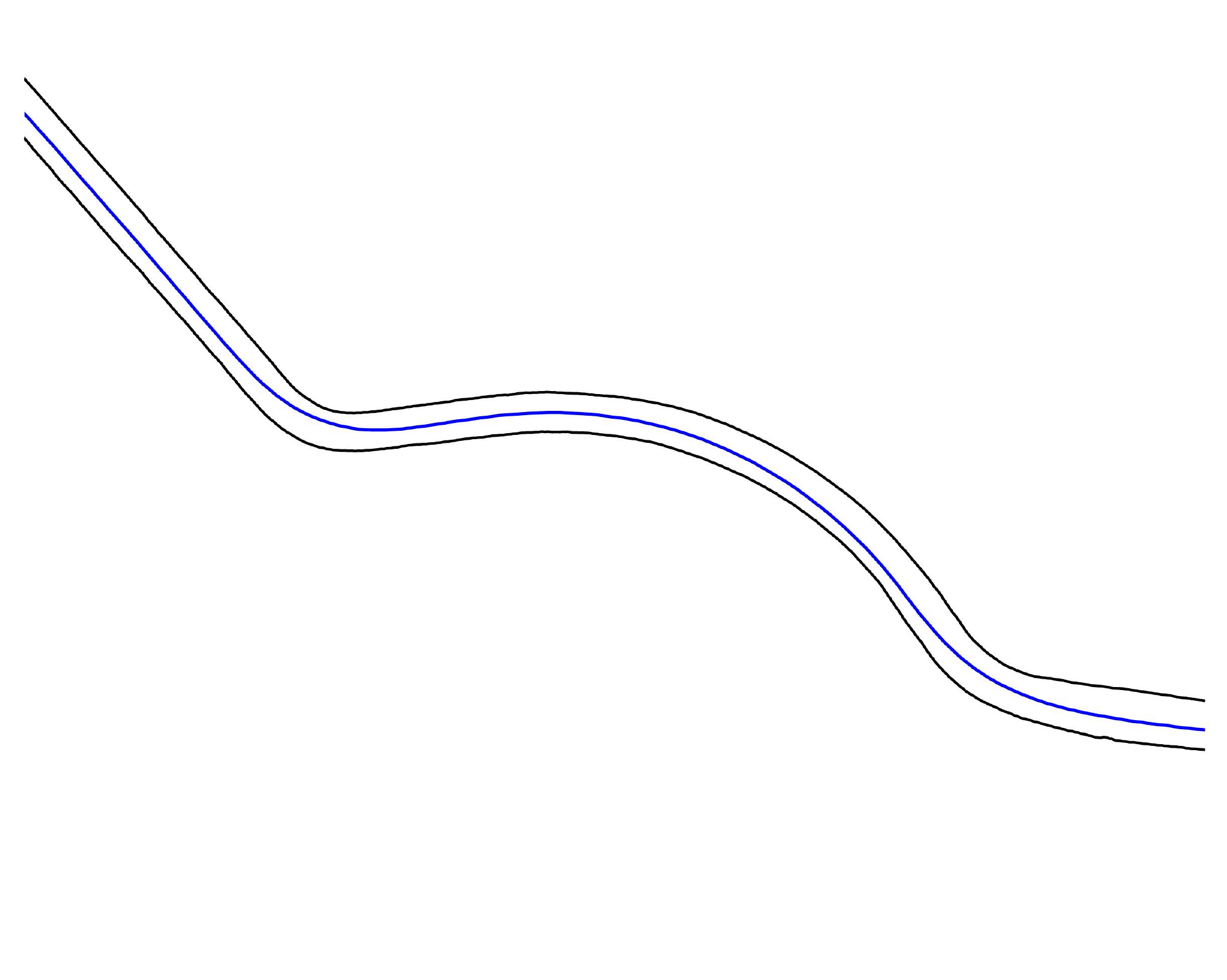}
}\hfill
\subfloat[Track B - Racing\label{b2}]{
    \includegraphics[width=0.47\linewidth]{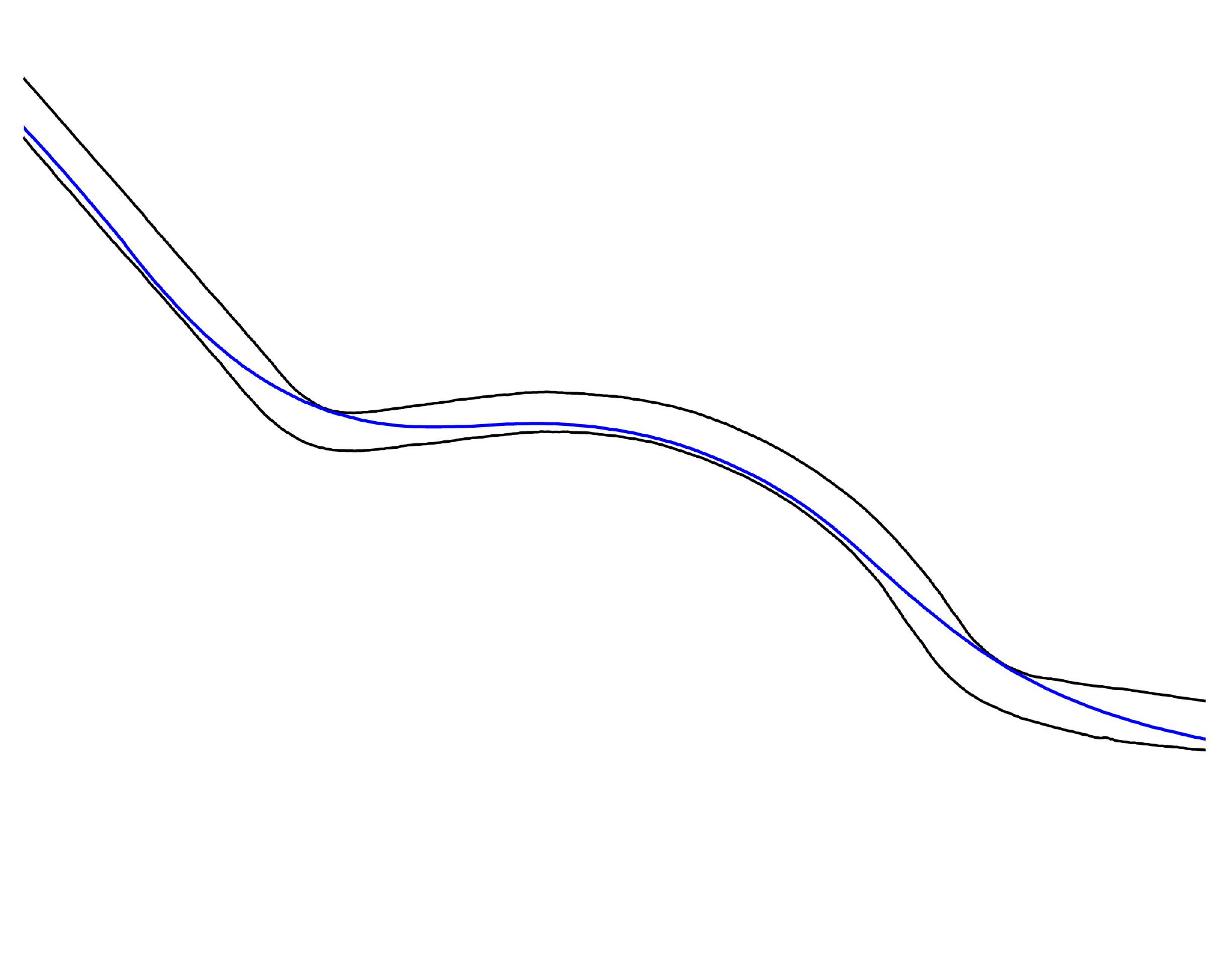}
}
\caption{Comparative example of the three driving styles recorded.}
\label{fig:driving_styles} 
\end{figure}
\par
\begin{figure*}[ht]%
\centering
\subfloat{
    \includegraphics[width=0.99\linewidth]{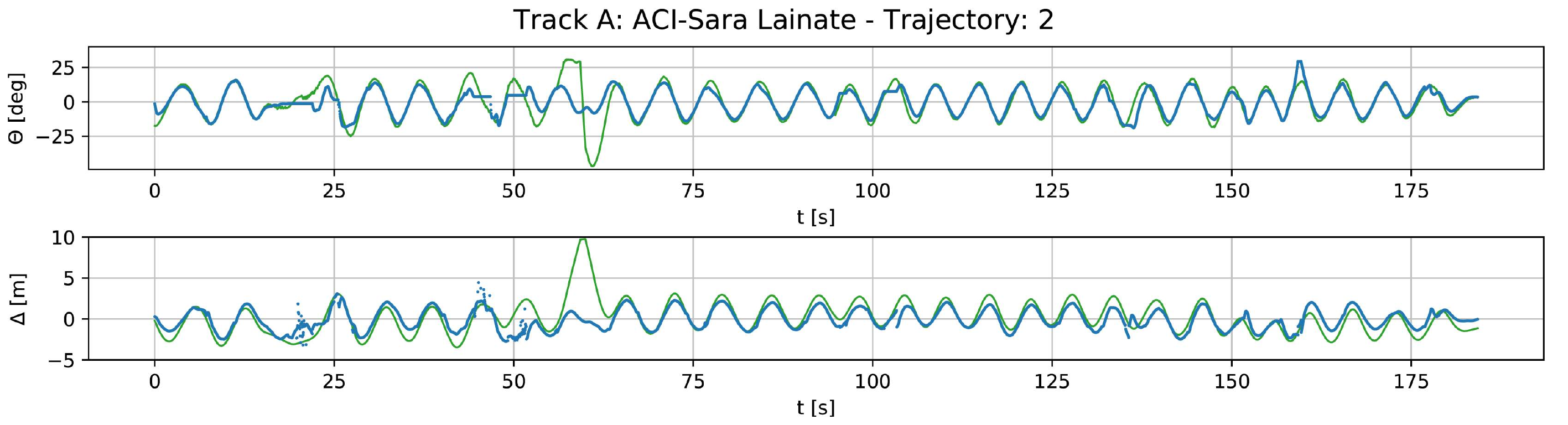}
}\\[5px]
\subfloat{
    \includegraphics[width=0.99\linewidth]{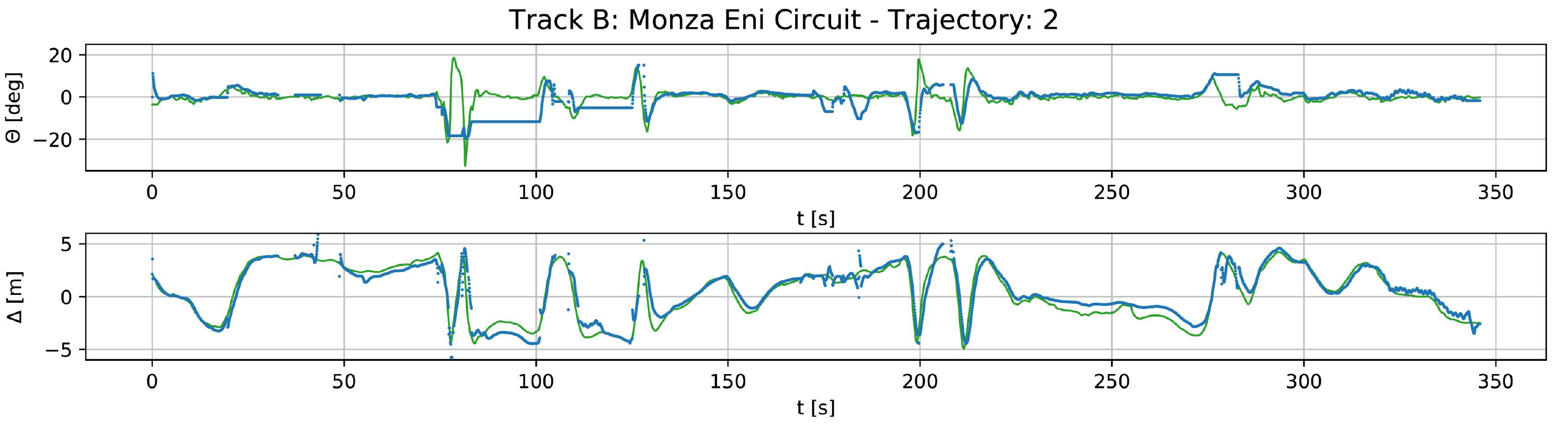}
}
\caption{Examples of the experimental results obtained in \textit{oscillating} and \textit{racing} driving styles, designed to stress the estimation of, respectively, heading (top) and lateral displacement (bottom). The graphs compare the estimates (in blue) with the respective ground truth (in green).}
\label{fig:results_plots} 
\end{figure*}

For the tests, we recorded multiple laps, with different speed, from $3\, m/s$ up to $15\, m/s$ and different driving styles. In particular, we considered three different trajectories (Fig.~\ref{fig:driving_styles}), one in the middle of the road, representing the optimal trajectory of the vehicle. Then, one oscillating, with significative heading changes, up to $40\degree$ to stress the limits of the algorithm, with particular focus to the heading value. Lastly, one on the racing line, often close to one side of the track and on the curbs, to better examine the computed lateral offset. Moreover, the recordings were performed in different weather scenarios, some on a sunny day, others with rain. This guarantees approximately one hour of recordings on two race track, one with length $1.5\, km$ and one with $5.8\, km$ extension, for a total of $30\, km$, with multiple driving styles and mixed atmospheric condition.
With those described, we evaluate the performance of our system in delivering the necessary information for the lateral control, i.e. the relative pose of the vehicle ($\Theta$, $\Delta$). The estimation is performed on four rides (two on each available track), covering three driving styles. To compare the results with the ground truth, we measure the mean absolute error reported on the entire experiment, considering only the frames where an estimate was available. A measure of the relative number of frames for which this happens ($Avail\ \%$) is also considered as an interesting metric. The results are reported in Table~\ref{tab:results_table}.
For further reference, the behavior of our estimates over time for
the most significant experiments is presented in Fig.~\ref{fig:results_plots}.
%
\begin{table}[t]
\renewcommand{\arraystretch}{1.3}
\centering
\caption{Results obtained with different driving styles}
\label{tab:results_table}
\begin{tabular}{@{}p{10em}rrr@{}}
	\hline
	& $ \mathit{MAE}_\Theta\,[\degree]$ & $\mathit{MAE}_\Delta \,[m]$ & $Avail\ \%$ \\
	\hline
	\multicolumn{1}{@{}l}{\emph{Driving style: centered}} \\
	\hspace{1em}Track A - Trajectory 1 & $2.892$ & $0.820$ & 99.71 \\
	\hspace{1em}Track B - Trajectory 1 & $1.642$ & $0.453$ & 99.91 \\
	\vspace{.01em} \\	
	\multicolumn{1}{@{}l}{\emph{Driving style: oscillating}} \\
	\hspace{1em}Track A - Trajectory 2 & $3.862$ & $0.946$ & 100.00  \\
	\vspace{.01em} \\	
	\multicolumn{1}{@{}l}{\emph{Driving style: racing}} \\
	\hspace{1em}Track B - Trajectory 2 & $3.120$ & $0.581$ & 95.92 \\
	\hline
\end{tabular}
\end{table}
\par
From these experiments, we observe how the system is able to provide an accurate estimate of the required data for lateral control, while maintaining a high availability.
Indeed, the errors registered for the lateral offset account for only $5 - 10 \%$ 
of the lane width, which lies between 9 to 12 meters for the tracks considered, while the errors in the heading, of about $3\degree$, are comparable to the ones experimentally obtained using a RTK GPS.
Furthermore, the error values remain considerably low also in non-optimal scenarios (Track A Trajectory 2, Track B Trajectory 2) where the vehicle follows a path considerably different from a normal driving style. 

\section{CONCLUSIONS}
In this paper, we propose a perception system relying on vision for the task of lateral control parameters estimation. This system is able to detect the lateral lines on the road and use them to estimate the lane centerline and relative pose of the vehicle in terms of heading and lateral displacement. As no benchmarking is publicly available, a custom dataset is collected and made openly available for future researches. The results obtained indicate that the proposed system can achieve high accuracy in different driving scenarios and weather conditions. The retrieved values are indeed comparable to the one calculated by a state of the art RTK GPS, while compensating for its shortcomings.


\bibliographystyle{IEEEtran}
\bibliography{IEEEabrv,ms}

\end{document}